\documentclass[runningheads]{llncs}

\usepackage{comment}
\usepackage{epsfig}
\usepackage{calc}
\usepackage{graphicx}
\usepackage{amsmath}
\usepackage{amssymb}
\usepackage[dvipsnames]{xcolor}
\usepackage{float}
\usepackage{subcaption}
\usepackage{booktabs}
\usepackage{xspace}
\usepackage{arydshln}
\usepackage{anyfontsize}
\usepackage{wrapfig}
\usepackage{multirow}
\usepackage{graphbox}
\usepackage{xspace}
\usepackage{hyperref}
\hypersetup{colorlinks=true}
\usepackage[
  width=122mm,left=12mm,paperwidth=146mm,height=193mm,
  top=12mm,paperheight=217mm]{geometry}

\newcommand{\smallcap}{\small}

\newcommand{\mypar}[1]{\paragraph{\textbf{#1}}}
\newcommand{\stimes}{{\mkern-2mu\times\mkern-2mu}}
\newcommand{\ioug}{{\textit{IoU}_\textit{g}}}

\makeatletter
\DeclareRobustCommand\onedot{\futurelet\@let@token\@onedot}
\def\@onedot{\ifx\@let@token.\else.\null\fi\xspace}
\def\eg{\emph{e.g}\onedot} 
\def\ie{\emph{i.e}\onedot} 
 
 \def\vs{\emph{vs}\onedot}

\makeatother

\newcommand{\pixv}[4]{
  \includegraphics[height=#4,align=c]{#1_fused}
  \hspace{-10pt}
  \bgroup
  \def\arraystretch{0}
  \setlength\tabcolsep{0pt}
  \begin{tabular}{c}
  \includegraphics[height=#4*\real{0.33333333}]{#1_rgb} \\
  \includegraphics[height=#4*\real{0.33333333}]{#1_rec#2} \\
  \includegraphics[height=#4*\real{0.33333333}]{#1_rec#3} \\
  \end{tabular}
  \egroup
}

\newcommand{\pixh}[4]{
  \bgroup
  \def\arraystretch{0}
  \setlength\tabcolsep{0mm}
  \begin{tabular}{ccc}
  \multicolumn{3}{c}{\includegraphics[width=#4]{#1_fused}} \\[1mm]
  \includegraphics[width=#4*\real{0.33333333}]{#1_rgb} &
  \includegraphics[width=#4*\real{0.33333333}]{#1_rec#2} &
  \includegraphics[width=#4*\real{0.33333333}]{#1_rec#3}
  \end{tabular}
  \egroup
}

\newcommand{\rott}[1]{\rotatebox[origin=lB]{90}{#1}}

\begin{document}
\pagestyle{headings}
\mainmatter

\title{
  CoReNet: Coherent 3D scene reconstruction from a single RGB image
}
\titlerunning{CoReNet: Coherent 3D scene reconstruction from a single RGB image}
\authorrunning{Stefan Popov \and Pablo Bauszat \and Vittorio Ferrari}
\author{
\hspace{-4mm}
Stefan Popov \qquad \quad\hspace{2mm}
Pablo Bauszat \qquad \quad\hspace{0mm}
Vittorio Ferrari \\
\vspace{1mm}\tt\scriptsize
spopov@google.com \qquad\hspace{2mm}
pablo.bauszat@gmail.com \qquad
vittoferrari@google.com
}
\institute{Google Research}
\maketitle

\begin{abstract}
Advances in deep learning techniques have allowed recent work to reconstruct the
shape of a single object given only one RBG image as input.
Building on common encoder-decoder architectures for this task, we propose
three extensions:
(1) ray-traced skip connections that propagate local 2D information to the
output 3D volume in a physically correct manner;
(2) a hybrid 3D volume representation that enables building translation
equivariant models, while at the same time encoding fine object details
without an excessive memory footprint;
(3) a reconstruction loss tailored to capture overall object geometry.
Furthermore, we adapt our model to address the harder task of reconstructing
multiple objects from a single image. We reconstruct all objects jointly in one
pass, producing a coherent reconstruction, where all objects live in a single
consistent 3D coordinate frame relative to the camera and they do not intersect
in 3D space. We also handle occlusions and resolve them by hallucinating the
missing object parts in the 3D volume.
We validate the impact of our contributions experimentally both on synthetic
data from ShapeNet as well as real images from Pix3D. Our method improves over
the state-of-the-art single-object methods on both datasets. Finally, we
evaluate performance quantitatively on multiple object reconstruction with
synthetic scenes assembled from ShapeNet objects.
\end{abstract}

\section {Introduction}
\label{sect:intro}

3D reconstruction is key to genuine scene understanding, going beyond 2D
analysis. Despite its importance, this task is exceptionally hard, especially in
its most general setting: from one RGB image as input. Advances in deep learning
techniques have allowed recent
work~\cite{mescheder19cvpr,chen19cvpr,wang18eccv,richter18cvpr,shin18cvpr,niu18cvpr,tulsiani17cvpr,choy16eccv,girdhar16eccv,wu16nips}
to reconstruct the shape of a single object in an image.

\begin{figure}[t]
  \centering
  \begingroup
  \setlength{\tabcolsep}{0pt}
  \renewcommand{\arraystretch}{0}
  \begin{tabular}{ccc}
  \includegraphics[width=.19\textwidth]{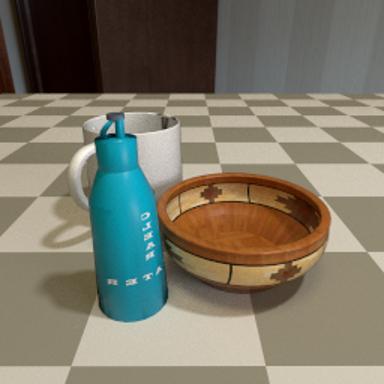}&
  \includegraphics[width=.19\textwidth]{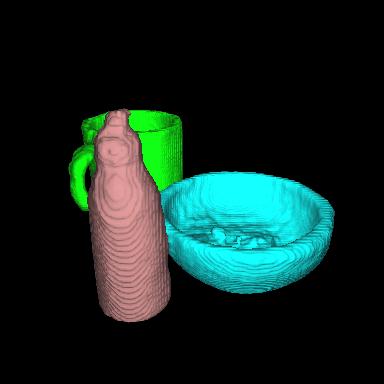}&
  \includegraphics[width=.19\textwidth]{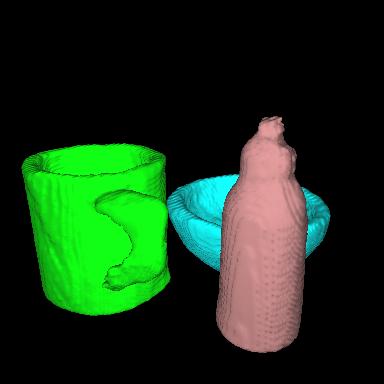}
  \\[1mm]
  \includegraphics[width=.19\textwidth]{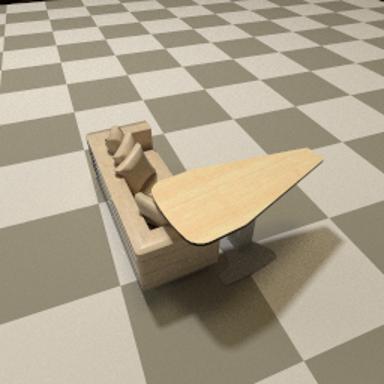}&
  \includegraphics[width=.19\textwidth]{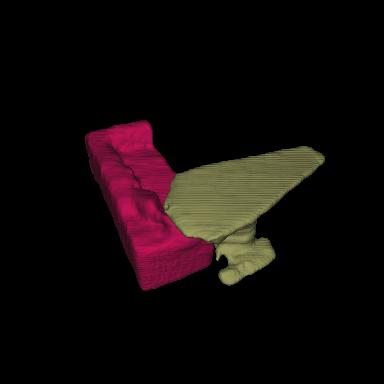}&
  \includegraphics[width=.19\textwidth]{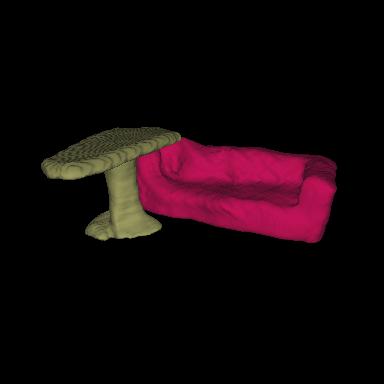}
  \end{tabular}
  \pixh{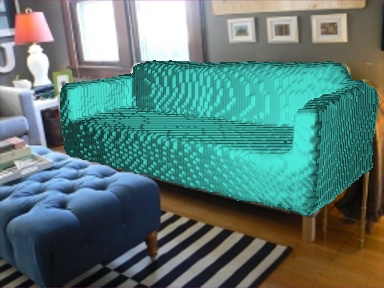}{3}{2}{46.5mm}
  \endgroup
  \caption{
  \small
  \smallcap
  3D reconstructions from a single RGB image, produced by our model.
  \textbf{Left:} Coherent reconstruction of multiple objects in a synthetic
  scene (shown from a view matching the input image, and another view). We
  reconstruct all objects in their correct spatial arrangement in a common
  coordinate frame, enforce space exclusion, and hallucinate occluded parts.
  \textbf{Right:} Reconstructing an object in a real-world scene. The top image
  shows the reconstruction overlaid on the RGB input. The bottom row shows the
  input next to two other views of the reconstruction.
  }
  \label{fig:teaser}
 \vspace{-1.5em}
\end{figure}

In this paper, we first propose several improvements for the task of
reconstructing a single object. As in~\cite{girdhar16eccv,xie19iccv}, we build
a neural network model which takes a RGB input image, encodes it and then
decodes it into a reconstruction of the full volume of the scene. We then
extract object meshes in a second stage. We extend this simple model with
three technical contributions:
(1)~Ray-traced skip connections as a way to propagate local 2D information to
the output 3D volume in a physically correct manner (sec.
\ref{sect:ray-traced-skip-connections}). They lead to sharp reconstruction
details because visible object parts can draw information directly from the
image;
(2)~A hybrid 3D volume representation that is both regular and implicit (sec.
\ref{sect:3d-volume-representation}). It enables building translation
equivariant 3D models using standard convolutional blocks, while at the same
time encoding fine object details without an excessive memory footprint.
Translation equivariance is important for our task, since objects can appear
anywhere in space;
(3)~A reconstruction training loss tailored to capture overall object
geometry, based on a generalization of the intersection-over-union metric
(IoU) (sec.~\ref{sect:iou-loss}).
Note that our model reconstructs objects at the pose (translation, rotation,
scale) seen from the camera, as opposed to a canonical pose in many
previous works~\cite{choy16eccv,girdhar16eccv,mescheder19cvpr,xie19iccv}.

We validate the impact of our contributions experimentally
on synthetic data from ShapeNet~\cite{chang16shapenet}
(sec.~\ref{sect:fgbg-reconstruction}) as well as real images from
Pix3D~\cite{sun18cvpr} (sec~\ref{sect:pix3d-results}).
The experiments demonstrate that
(1) our proposed ray-traced skip connections and IoU loss improve reconstruction
performance considerably;
(2) our proposed hybrid volume representation enables to reconstruct at
resolutions higher than the one used during training;
(3) our method improves over the state-of-the-art single-object 3D
reconstruction methods on both ShapeNet and Pix3D datasets.

In the second part of this paper, we address the harder task of reconstructing
scenes consisting of spatial arrangements of multiple objects. In addition to
the shape of individual objects at their depicted pose, we also predict the
semantic class of each object. We focus on coherent reconstruction in this
scenario, where we want to
(1)~reconstruct all objects and the camera at their correct relative pose in a
single consistent 3D coordinate frame,
(2)~detect occlusions and resolve them fully, hallucinating missing parts (\eg a
chair behind a table),
and (3)~ensure that each point in the output 3D space is occupied by at most
one object (space exclusion constraint).
We achieve this through a relatively simple modification of our single-object
pipeline. We predict a probability distribution over semantic classes at each
point in the output 3D space and we make the final mesh extraction step aware
of this.

The technical contributions mentioned above for the single-object case are
even more relevant for reconstructing scenes containing multiple objects.
Ray-traced skip connections allow the model to propagate occlusion boundaries
and object contact points detected on the 2D image into 3D, and to also
understand the depth relations among objects locally. The IoU loss teaches our
model to output compact object reconstructions that do not overlap in 3D
space. The hybrid volume representation provides a fine discretization
resolution, which can compensate for the smaller fraction of the scene volume
allocated to each object in comparison to the single object case.

We experimentally study our method's performance on multiple object
reconstruction with synthetic scenes assembled from ShapeNet objects (sec.
\ref{sect:multiclass-reconstruction}). We validate again the impact of our
technical contributions, and study the effect of the degree of object
occlusion, distance to the camera, number of objects in the scene, and their
semantic classes. We observe that ray-traced skip connections and the IoU loss
bring larger improvements than in the single object case. We show that our
model can handle multiple object scenes well, losing only a fraction of its
performance compared to the single object case.

Finally, we study the effect of image realism on reconstruction performance
both in the single-object (sec.~\ref{sect:fgbg-reconstruction}) and multi-object
(sec. \ref{sect:multiclass-reconstruction}) cases.
We render our  images with either
(1) local illumination against uniform background like most previous
works~\cite{mescheder19cvpr,wang18eccv,richter18cvpr,tulsiani17cvpr,choy16eccv,girdhar16eccv}
or
(2) a physically-based engine~\cite{pharr16book}, adding global
illumination effects, such as shadows and reflections, non-trivial background,
and complex lighting from an environment map and finite extent light sources.
We publicly release these images, our models, and scene
layouts~\cite{popov20corenet_project_url}.
\vspace{-1em}
\section{Related work}

\mypar{Single object reconstruction.}
In the last few years there has been a surge of methods for reconstructing the
3D shape of one object from a single RGB image. Many of
them~\cite{choy16eccv,girdhar16eccv,wu16nips,xie19iccv,xie20ijcv} employ voxel
grids in their internal representation, as they can be handled naturally by
convolutional neural networks.
Some works have tried to go beyond voxels:
(1) by using a differentiable voxels-to-mesh operation~\cite{liao18cvpr};
(2) by producing multiple depth-maps and/or silhouettes from fixed viewpoints
that can be subsequently
fused~\cite{soltani17cvpr,shin18cvpr,richter18cvpr,yao20cvpr};
(3) by operating on point clouds~\cite{fan17cvpr,mandikal18bmvc}, cuboidal
primitives~\cite{tulsiani17cvpr,niu18cvpr}, and even directly on
meshes~\cite{wang18eccv,chen20cvpr}.
A recent class of methods~\cite{park19cvpr,mescheder19cvpr,chen19cvpr} use a
continuous volume representation through implicit functions. The model
receives a query 3D point as part of its input and returns the occupancy at
that point.

We build on principles from these works and design a new type of hybrid
representation that is both regular like voxels and continuous like implicit
functions (sec.~\ref{sect:3d-volume-representation}). We also address more
complex reconstruction tasks: we reconstruct objects in the pose as depicted
in the image and we also tackle scenes with multiple objects, predicting the
semantic class of each object. Finally, we experiment with different levels of
rendering realism.

\mypar{Multi-object reconstruction.}
IM2CAD~\cite{izadinia17cvpr} places multiple CAD models from a database in their
appropriate position in the scene depicted in the input image. It only
reconstructs the pose of the objects and copies over their whole CAD models,
without trying to reconstruct their particular 3D shapes as they appear in the
input image.
3D-RCNN~\cite{kundu18cvpr} learns a per-class linear shape basis from a training
dataset of 3D models. It then uses a render-and-compare approach to fit the
coefficients of this basis to objects detected in the test image.
This method only outputs 3D shapes that lie on a simple linear subspace
spanning the training samples. Instead our model can output arbitrary shapes,
and the mapping between image appearance and shape is more complex as it is
modeled by a deep neural network.
Tulsiani et. al.~\cite{tulsiani18cvpr} first detects object
proposals~\cite{zitnick14eccv} and then reconstructs a pose and a voxel grid for
each, based on local features for the proposal and a global image descriptor.
Mesh-RCNN~\cite{gkioxari19iccv} extends Mask-RCNN~\cite{he17iccv} to predict a
3D mesh for each detected object in an image. It tries to predict the objects
positions in the image plane correctly, but it cannot resolve the fundamental
scale/depth ambiguity along the Z-axis.

All four methods~\cite{izadinia17cvpr,kundu18cvpr,tulsiani18cvpr,gkioxari19iccv}
first detect objects in the 2D image, and then reconstruct their 3D shapes
independently. Instead, we reconstruct all objects jointly and without relying
on a detection stage. This allows us to enforce space exclusion constraints and
thus produce a globally coherent reconstruction.

The concurrent work~\cite{nie20cvpr} predicts the 3D pose of all objects jointly
(after 2D object detection). Yet, it still reconstructs their 3D shape
independently, and so the reconstructions might overlap in 3D space. Moreover,
in contrast to~\cite{nie20cvpr,tulsiani18cvpr,izadinia17cvpr,kundu18cvpr} our
method is simpler as it sidesteps the need to explicitly predict per-object
poses, and instead directly outputs a joint coherent reconstruction.

Importantly, none of these
works~\cite{izadinia17cvpr,kundu18cvpr,gkioxari19iccv,nie20cvpr} offers true
quantitative evaluation of 3D shape reconstruction on multiple object scenes.
One of the main reasons for this is the lack of datasets with complete and
correct ground truth data. One exception is~\cite{tulsiani18cvpr} by evaluating
on SunCG~\cite{song16cvpr}, which is now banned.
In contrast, we evaluate our method fully, including the 3D shape of multiple
objects in the same image. To enable this, we create two new datasets of scenes
assembled from pairs and triplets of ShapeNet objects, and we report performance
with a full scene evaluation metric (sec.~\ref{sect:multiclass-reconstruction}).

Finally, several works tackle multiple object reconstruction from an RGB-D
image~\cite{nicastro19iccv,song17cvpr}, exploiting the extra information that
depth sensors provides.

\mypar{Neural scene representations.} Recent
works~\cite{saito19iccv,sitzmann19cvpr,nguyen-phuoc18nips,nguyen-phuoc19iccv,sitzmann19nips,lombardi19tog}
on neural scene representations and neural rendering extract latent
representations of the scene geometry from images and share similar insights to
ours. In particular, \cite{kar17nips,tung19cvpr} use unprojection, a technique
to accumulate latent scene information from multiple views, related to our
ray-traced skip connections. Others~\cite{saito19iccv,sitzmann19nips} can also
reconstruct (single-object) geometry from one RGB image.

\section{Proposed approach}
\label{sect:proposed-approach}

\begin{figure}[t]
  \centering
  \includegraphics[width=.3\textwidth]{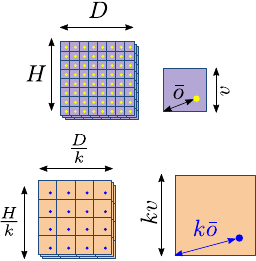}
  \hspace{.05\textwidth}
  \includegraphics[width=.63\textwidth]{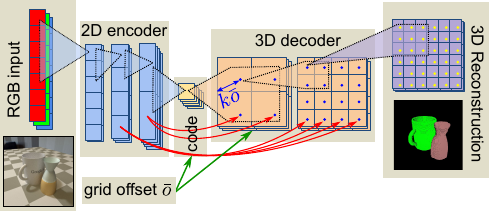}
  \caption{
  \small
  \smallcap
  \textbf{Left:} 2D slice of the output grid (yellow points) and a decoder
  layer grid (blue points). The output grid is offset by $\bar{o}$ from the
  origin. The decoder grid, which has $k$ times lower resolution, by
  $k\bar{o}$.
  \textbf{Right:} Side-cut of our model's architecture. Ray-traced skip
  connections (red) propagate data from the encoder to the decoder, $\bar{o}$
  is appended to the channels of select decoder layers (green).
  }
 \vspace{-1.5em}
  \label{fig:model-general-arhitecture}
\end{figure}

For simplicity and compactness of exposition, we present directly
our full method, which can reconstruct multiple objects in the same image. Our
reconstruction pipeline takes a single RGB image as input and outputs a set of
meshes -- one for each object in the scene. It is trained to jointly predict the
object shapes, their pose relative to the camera, and their class label.

At the core of our pipeline is a neural network model that receives a single
RGB image and a set of volume query points as input and outputs a probability
distribution over $C$ possible classes at each of these points. One of the
classes is \emph{void} (\ie empty space), while the rest are object classes,
such as chair and table. Predicting normalized distributions creates
competition between the classes and forces the model to learn about space
exclusion. For single object models, we use two classes (foreground and
\emph{void}, $C=2$).

To create a mesh representation, we first reconstruct a fine discretization of
the output volume (sec.~\ref{sect:mesh-reconstruction}). We query the model
repeatedly, at different locations in 3D space, and we integrate the obtained
outputs. We then apply marching cubes~\cite{lewiner03jgtools} over the
discretization, in a way that enforces the space exclusion constraint. We
jitter the query points randomly during training. For single object models, we
treat all meshes as parts of one single output object.

\subsection{3D volume representation}
\label{sect:3d-volume-representation}

We want our model to reconstruct the large 3D scene volume at a fine
spatial resolution, so we can capture geometric details of individual objects,
but without an excessive memory footprint. We also want it to be
\emph{translation equivariant}: if the model sees \eg chairs only in one
corner of the scene during training, it should still be able to reconstruct
chairs elsewhere. This is especially important in a multi-object
scenario, where objects can appear anywhere in space.

A recent class of models~\cite{mescheder19cvpr,chen19cvpr,park19cvpr} address
our first requirement through an implicit volume representation. They input a
compact code, describing the volume contents, and a query point. They output the
occupancy of the volume at that point. These models can be used for
reconstruction by conditioning on a code extracted from an image, but are not
translation equivariant by design.

Models based on a voxel grid representation~\cite{choy16eccv,girdhar16eccv}
are convolutional in nature, and so address our translation equivariance
requirement, but require excessive memory to represent large scenes at fine
resolutions (cubic in the number of voxels per dimension).

We address both requirements with a new hybrid volume representation and a
model architecture based on it. Our model produces a multinomial distribution
over the $C$ possible classes on a regular grid of points. The structure of
the grid is fixed (\ie fixed resolution $W\stimes H\stimes D$ and distance
between points $v$), but we allow the grid to be placed at an arbitrary
spatial offset $\bar{o}$, smaller than $v$
(fig.~\ref{fig:model-general-arhitecture}). The offset value is an input to
our model, which then enables fine-resolution reconstruction (see below).

This representation combines the best of voxel grids and implicit volumes. The
regular grid structure allows to build a fully convolutional model that is
translation equivariant by design, using only standard 3D convolution building
blocks. The variable grid offset allows to reconstruct regular samplings of
the output volume at any desired resolution (multiple of the model grid's
resolution), while keeping the model memory footprint constant.
To do this, we call our model repeatedly with different appropriately chosen
grid offsets during inference (sec. \ref{sect:mesh-reconstruction}) and
integrate the results into a single, consistent, high-resolution output. We
sample the full output volume with random grid offsets $\bar{o}$ during
training.

\subsection{Core model architecture}
\label{sect:core-model-architecture}

We construct our model on top of an encoder-decoder skeleton (fig.
\ref{fig:model-general-arhitecture}). A custom decoder transforms the output of
a standard ResNet-50~\cite{he16cvpr} encoder into a $W \stimes H \stimes D
\stimes C$ output tensor -- a probability distribution over the $C$ possible
classes for each point in the output grid.
The decoder operations alternate between upscaling, using transposed 3D
convolutions with stride larger than 1, and data mixing while preserving
resolution, using 3D convolutions with stride 1.

We condition the decoder on the grid offset $\bar{o}$. We further create
\emph{ray-traced skip connections} that propagate information from the encoder
to the decoder layers in a physically accurate manner in sec.
\ref{sect:ray-traced-skip-connections}.
We inject $\bar{o}$ and the ray-traced skip connections before select data
mixing operations.

\subsection{Ray traced skip connections}
\label{sect:ray-traced-skip-connections}

\begin{wrapfigure}{r}{.5\linewidth}
  \vspace{-4em}
  \centering
  \includegraphics[width=\linewidth]{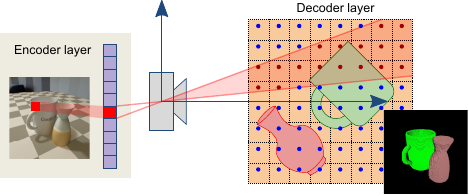}
  \vspace{-2em}
  \caption{
  \small
  \smallcap
  Pixels in the 2D encoder embed local image information, which ray-traced
  skip connections propagate to all 3D decoder grid points in the
  corresponding frustum.}
  \vspace{-2.5em}
  \label{fig:rt-skip-conn}
\end{wrapfigure}
So far we relied purely on the encoder to learn how to reverse the physical
process that converts a 3D scene into a 2D image. This process is well
understood however~\cite{hartley00,pharr16book} and many of its elements have
been formalized mathematically.
We propose to inject knowledge about it into
the model, by connecting each pixel in the input image to its corresponding
frustum in the output volume (fig. \ref{fig:rt-skip-conn}).

We assume for now that the camera parameters are known. We can compute the 2D
projection on the image plane of any point in the 3D output volume and we use
this to build ray-traced skip connections. We choose a source 2D encoder layer
and a target 3D decoder layer. We treat the $W_e \stimes H_e \stimes C_e$
encoder layer as a $W_e \stimes H_e$ image with $C_e$ channels, taken by our
camera. We treat the $W_d \stimes H_d \stimes D_d \stimes C_d$ decoder layer
as a $W_d \stimes H_d \stimes D_d$ grid of points. We project the decoder
points onto the encoder image, then sample it at the resulting
2D coordinates, and finally carry the sampled data over to the 3D decoder.
This creates skip connections in the form of rays that start at the camera
image plane, pass through the camera pinhole and end at the decoder grid point
(fig. \ref{fig:rt-skip-conn}).
We connect several of the decoder layers to the encoder in this manner,
reducing the channel count beforehand to $0.75 \cdot C_d$ by using $1 \stimes
1$ convolutions.

\mypar{Decoder grid offset.}
An important detail is how to choose the parameters of the decoder layer's grid.
The resolution is determined by the layer itself (\ie $W_d
\stimes H_d \stimes D_d$). It has $k$ times lower resolution than the
output grid (by design). We choose $v_d=kv$ for distance between the grid
points and $\bar{o_d}=k\bar{o}$ for grid offset
(fig.~\ref{fig:model-general-arhitecture}).
This makes the decoder grid occupy the same space as the final output grid and
respond to changes in the offset $\bar{o}$ in a similar way. In turn, this
aids implicit volume reconstruction in sec. \ref{sect:mesh-reconstruction}
with an additional parallax effect.

\mypar{Obtaining camera parameters.}
Ray-traced skip connections rely on known camera parameters. In practice, the
intrinsic parameters are often known. For individual images, they can be
deduced from the associated metadata (\eg EXIF in JPEGs). For 3D datasets such
as Pix3D~\cite{sun18cvpr} and Matterport3D~\cite{chang173dv} they are usually
provided.
When not available, we can assume default intrinsic parameters, leading to
still plausible 3D reconstructions (\eg correct relative
proportions but wrong global object scale).
The extrinsic parameters in contrast are usually unknown. We compensate for
this by reconstructing relative to the \emph{camera} rather than in world
space, resulting in an identity extrinsic camera matrix.

\subsection{IoU training loss}
\label{sect:iou-loss}

The output space of our model is a multinomial distribution over
the $C$ possible classes (including {\em void}), for each point in 3D space.
This is analog to multi-class recognition in 2D computer vision and hence we
could borrow the categorical cross-entropy loss common in those
works~\cite{he16cvpr,he17iccv}.
In our case, most space in the output volume in empty, which
leads to most predicted points having the {\em void} label. Moreover,
as only one object can occupy a given point in space, then all but one of the
$C$ values at a point will be $0$. This leads to even more sparsity.
A better loss, designed to deal with extreme class imbalance, is the
\emph{focal loss}~\cite{lin17iccv}.

Both categorical cross-entropy and the focal loss treat points as a batch of
independent examples and average the individual losses. They are not well
suited for 3D reconstruction, as we care more about overall object geometry,
not independent points. The 3D IoU metric is better suited to capture this,
which inspired us to create a new \emph{IoU loss}, specifically aiming to
minimize it. Similar losses have been successfully applied to 2D image
segmentation problems~\cite{sudre17dlmia,berman18cvpr}.

We generalize IoU, with support for continuous values and multiple classes:
\noindent\begin{minipage}{\linewidth}
\vspace{.1em}
\fontsize{7}{9}\selectfont
\begin{equation}
\textit{IoU}_\textit{g}(g, p) = \frac{
  \sum\limits_{i \in G}{\sum\limits_{c=1}^{C-1}{
    \min(g_{ic}, p_{ic})\cdot\mu(g_{ic})}
}}{
  \sum\limits_{i \in G}{\sum\limits_{c=1}^{C-1}{
    \max(g_{ic}, p_{ic})\cdot \mu(g_{ic})}
  }}
, \ \
\mu(g_{ic}) =
\begin{cases}
  1 &, \textrm{if } g_{ic} = 1 \\
  \frac{1}{C-1} &, \textrm{if } g_{ic} = 0
\end{cases}
\label{eq:ioug}
\end{equation}
\vspace{.1em}
\end{minipage}
where $i$ loops over the points in the grid, $c$ -- over the $C-1$ non-void
classes, $g_{ic} \in \{0,1\}$ is the one-hot encoding of the ground truth
label, indicating whether point $i$ belongs to class $c$, and $p_{ic}
\in [0,1]$ is the predicted probability. $\mu(g_{ic})$ balances for the
sparsity due to multiple classes, as $C-1$ values in the ground truth one-hot
encoding will be 0.

With two classes (\ie $C=2$) and binary values for $p$ and $g$,
$\ioug$ is equivalent to the intersection-over-union measure. The
$\max$ operator acts like logical {\em and}, $\min$ like logical {\em or}, and
$\mu(g_{ic})$ is always one.
In the case where there is a single object class to be reconstructed we use
$1-\ioug$ as a loss (sec.~\ref{sect:fgbg-reconstruction}). With multiple
objects, we combine $\ioug$ with categorical cross entropy into a product
(sec.~\ref{sect:multiclass-reconstruction})

\subsection{Mesh reconstruction}
\label{sect:mesh-reconstruction}
Our end-goal is to extract a set of meshes that represent the surface of the
objects in the scene. To do this, we first reconstruct an arbitrary fine
discretization of the volume, with a resolution that is an integer multiple
$n$ of the model's output resolution. We call the model $n^3$ times, each time
with a different offset
$\bar{o} \in \left\{
  \frac{0 + 0.5}{n} v,
  \frac{1 + 0.5}{n} v,
  ...,
  \frac{n-1 + 0.5}{n} v \right\}^3$
and we interleave the obtained grid
values. The result is a $nW \stimes nH \stimes nD \stimes C$ discretization
of the volume.

We then extract meshes. We break the discretization into $C$ slices of shape
$nW \stimes nH \stimes nD$, one for each class. We run marching
cubes~\cite{lewiner03jgtools} with threshold $0.5$ on each slice independently
and we output meshes, except for the slice corresponding to the \emph{void}
class. The $0.5$ threshold enforces space exclusion, since at most one value
in a probability distribution can be be larger than $0.5$.

\section{Experiments}
\label{sect:experiments}

\begin{figure}[t!]
  \centering
  \begingroup
  \setlength{\tabcolsep}{1pt}
  \renewcommand{\arraystretch}{0.5}
  \begin{tabular}{cc}
  \includegraphics[width=.129\textwidth]{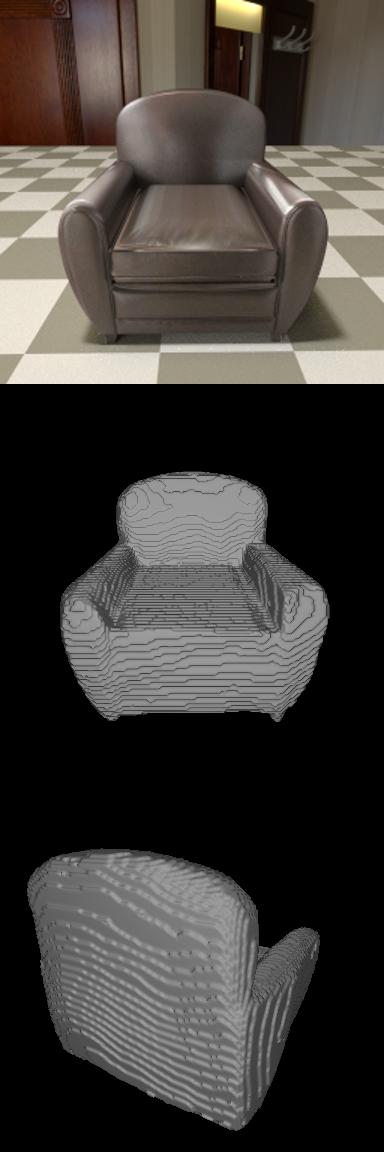}
  \includegraphics[width=.129\textwidth]{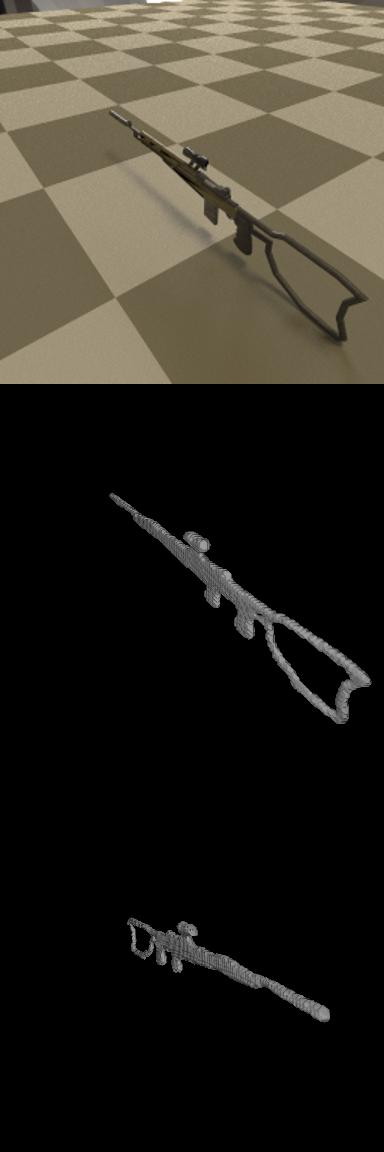}
  \includegraphics[width=.129\textwidth]{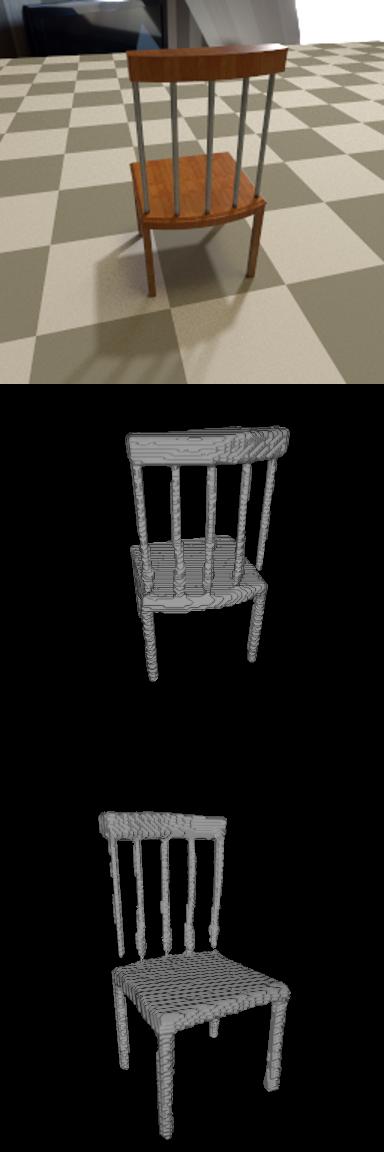}
  \includegraphics[width=.129\textwidth]{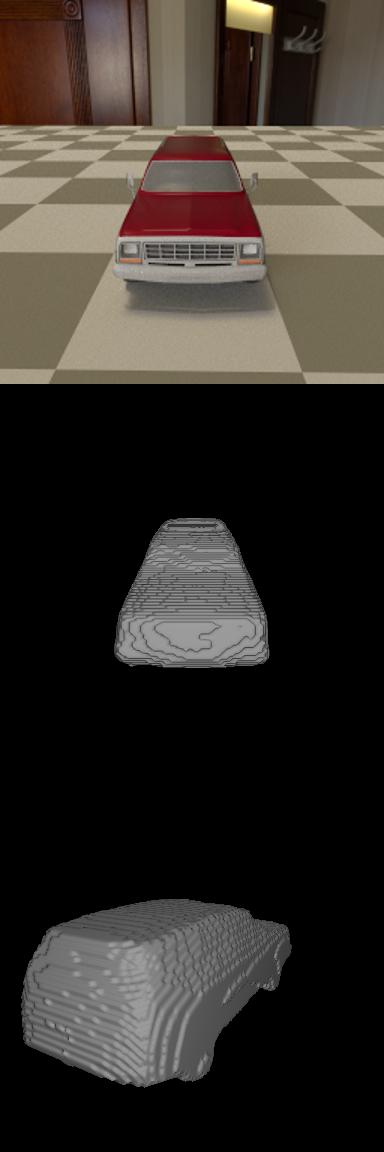}
  \includegraphics[width=.129\textwidth,trim={0 92 0 0},clip]{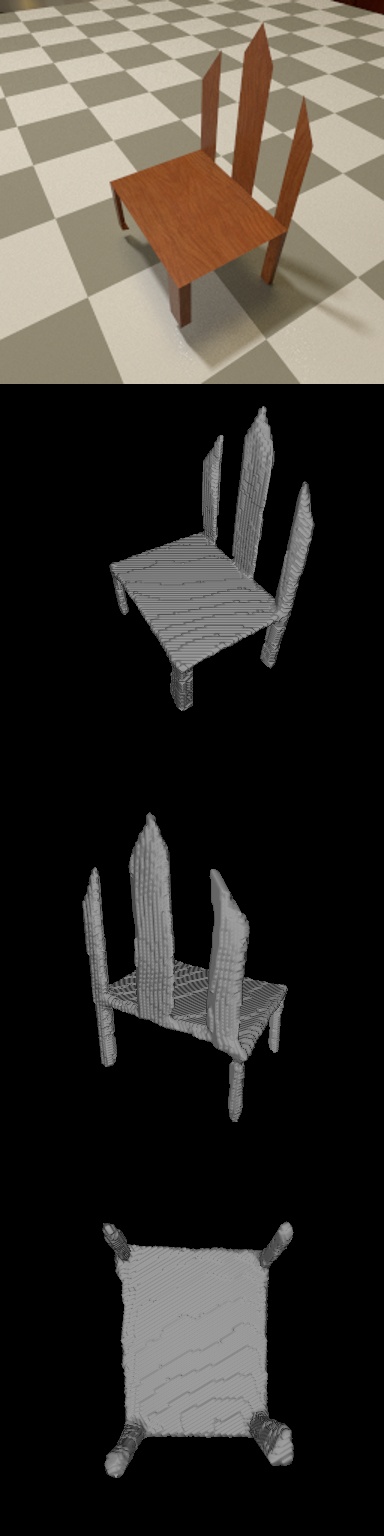}
  \hspace{1em}
  \includegraphics[width=.129\textwidth]{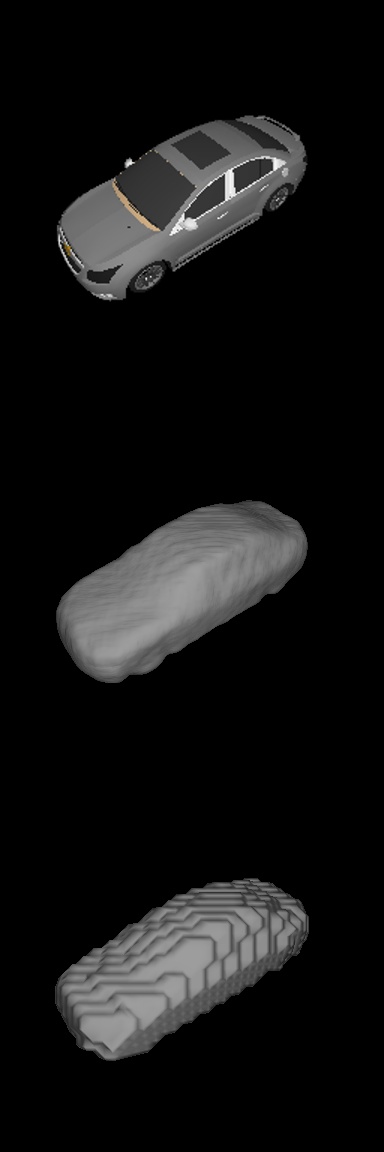}
  \includegraphics[width=.129\textwidth]{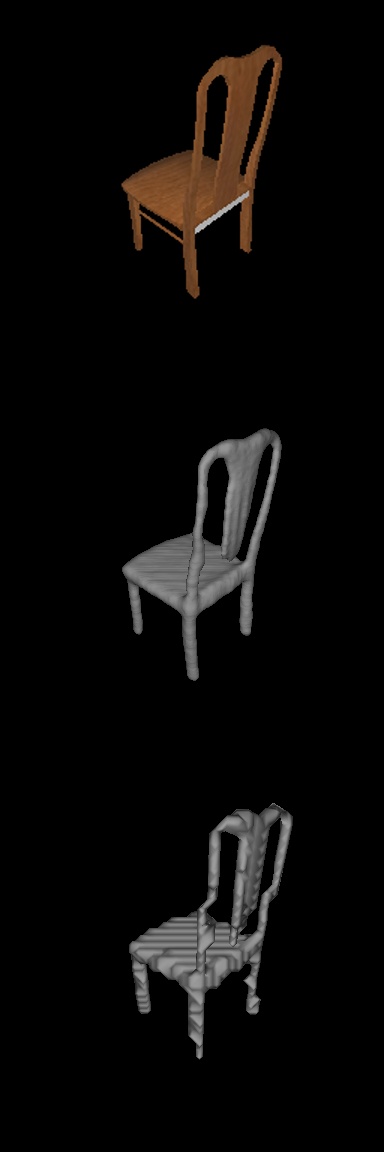}
  \end{tabular}
  \endgroup
   \vspace{-.5em}
  \caption{
    \small
    \smallcap
    Single object experiments (sec.~\ref{sect:single-object-performance}).
    \textbf{Left:} Scenes reconstructed by $h_7$, shown from two different
    viewpoints. Our model handles thin structures and hallucinates invisible
    back-facing object parts.
    \textbf{Right:} Scenes reconstructed by $y_1$. Despite the low resolution
    of $y_1$ ($32^3$, second row), we reconstruct high-quality meshes (first
    row) by sampling $y_1$ with $4^3$ grid offsets (see
    sec.~\ref{sect:proposed-approach}).
  }
  \label{fig:single-object-reconstructions}
  \label{fig:superres-results}
  \vspace{-1.0em}
\end{figure}

We first present experiments on single object reconstruction on synthetic
images from ShapeNet~\cite{chang16shapenet}
(sec.~\ref{sect:fgbg-reconstruction}) and on real images from
Pix3D~\cite{sun18cvpr} (sec.~\ref{sect:pix3d-results}). Then we evaluate
performance on multiple object reconstruction in
sec.~\ref{sect:multiclass-reconstruction}.

\subsection{Single object reconstruction on ShapeNet}
\label{sect:fgbg-reconstruction}
\label{sect:single-object-performance}

\begin{wrapfigure}{r}{0.24\linewidth}
 \vspace{-2.8em}
  \centering
  \begin{tabular}{cc}
  \includegraphics[width=.465\linewidth]{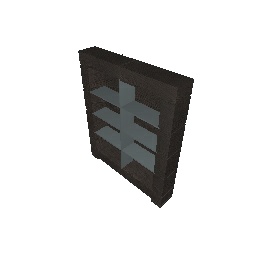}&
  \includegraphics[width=.465\linewidth]{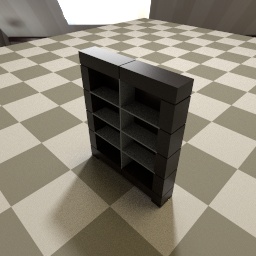}
  \\
  \includegraphics[width=.465\linewidth]{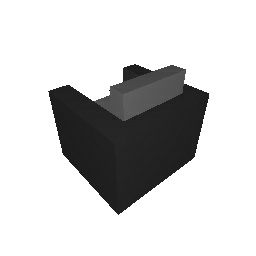}&
  \includegraphics[width=.465\linewidth]{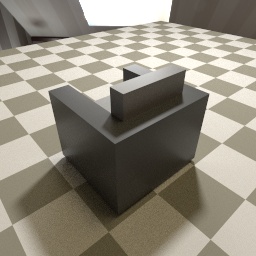}
  \end{tabular}
 \vspace{-2.4em}
\end{wrapfigure}

\mypar{Dataset.}
We use ShapeNet~\cite{chang16shapenet}, following the setting
of~\cite{choy16eccv}. We consider the same 13 classes, train on 80\% of the
object instances and test on 20\% (the official ShapeNet-trainval and
ShapeNet-test splits). We normalize and center each object in the unit cube and
render it from 24 random viewpoints, with two levels of photorealism (see inset
figure on the right): \emph{low realism}, using local illumination on a uniform
background, with no secondary effects such as shadows and reflections; and
\emph{high realism}, with full global illumination using PBRT's
renderer~\cite{pharr16book}, against an environment map background, and with a
ground plane. The low realism setting is equivalent to what was used in previous
works~\cite{mescheder19cvpr,wang18eccv,choy16eccv}.

\mypar{Default settings.}
Unless specified otherwise, we train and evaluate at the same grid resolution
($128^3$), and we use the same camera parameters in all scenes. We evaluate
intersection-over-union as a volumetric metric, reporting mean over the classes
(mIoU) as well as the global mean over all object instances.
We also evaluate the predicted meshes with the
F@1\%-score~\cite{tatarchenko19cvpr} as a surface metric. As commonly
done~\cite{park19cvpr,mescheder19cvpr,chen19cvpr}, we pre-process the
ground-truth meshes to make them watertight and to remove hidden and duplicate
surfaces. We sample all meshes uniformly with 100K points, then compute F-score
for each class, and finally report the average over classes.

\mypar{Reconstruction performance.}
We report the effect of hyper parameters on performance in
table~\ref{tab:single-object-hyper-param}(a) and show example reconstructions in
fig.~\ref{fig:single-object-reconstructions}.
Ray-traced skip connections improve mIoU by about $5\%$ and F@1\% by 10\%,
in conjunction with any loss. Our IoU loss performs best, followed by focal and
categorical cross entropy (Xent).
Somewhat surprisingly, results are slightly better on images with high realism,
even though they are visually more complex. Shadows and reflections might be
providing additional reconstruction cues in this case. Our best model for low
realism images is $h_5$ and for high realism it is $h_7$.

\begin{table}[t!]
  \fontsize{9}{11}
  \selectfont
  \begin{tabular}{c|ccc|cc|cc}
  & \textbf{skip}&  &  \textbf{rea-}& \multicolumn{2}{c|}{\textbf{IoU}} & \\
  \textbf{id} & \textbf{conn.}& \textbf{loss} & \textbf{lism} & \textbf{mean} &
  \textbf{glob.} & \textbf{F@1\% } \\
  \hline
    $h_1$ & No  & focal & low & 50.8 & 52.0 & 45.0 \\
    $h_2$ & No  & IoU & low & 53.0 & 53.9  & 47.8 \\
  \hline
    $h_3$ & Yes & Xent & low & 54.1 & 55.2 & 52.9 \\
    $h_4$ & Yes & focal & low & 56.6 & 57.5 & 54.4 \\
    $h_5$ & Yes & IoU & low & \textbf{57.9} & 58.7 & \textbf{57.5} \\
  \hline
    $h_6$ & Yes & Focal & high & 58.1 & 58.4  & 57.3 \\
    $h_7$ & Yes & IoU & high & \textbf{59.1} & 59.3 & \textbf{59.5} \\
  \end{tabular}
  \begin{tabular}{c|cccc|cc}
  & & \textbf{skip} & & \textbf{rea-} & \multicolumn{2}{c}{\textbf{IoU}} \\
  \textbf{id}& \textbf{data} & \textbf{conn.} & \textbf{loss} & \textbf{lism} &
  \textbf{mean} & \textbf{glob.} \\
  \hline
$m_1$ & pairs & no & focal & high & 34.9 & 46.4 \\
$m_2$ & pairs & no & IoU  &high &  33.1 & 43.4 \\
  \hline
$m_3$ & pairs & yes & focal & low & 40.4 & 49.7 \\
$m_4$ & pairs & yes & IoU & low & 41.8 & 50.6 \\
  \hline
$m_5$ & pairs & yes & Xent & high & 30.0 & 43.5 \\
$m_6$ & pairs & yes & focal & high & 42.7 & 52.4 \\
$m_7$ & pairs & yes & IoU & high & \textbf{43.1} & 52.7 \\
  \hline
$m_8$ & tripl. & yes & focal & high & 43.0 & 49.1 \\
$m_9$ & tripl. & yes & IoU & high & \textbf{43.9} & 49.8 \\
  \hline
$m_{10}$ & single & yes & focal & high & 43.4 & 53.9 \\
$m_{11}$ & single & yes & IoU & high & 46.9 & 56.4 \\
  \end{tabular}

  \small
  \centering
  \vspace{.5em}
  \caption{
  \small
  \smallcap
  Reconstruction performance in \% for \textbf{(a)} our single object
  experiments on the left, and \textbf{(b)} our multiple object experiments on
  the right.
  }
  \label{tab:single-object-hyper-param}
  \label{tab:multi-object-results}
  \vspace{-2em}
\end{table}

\mypar{Comparison to state-of-the-art.}
We compare our models to state-of-the art single object reconstruction
methods~\cite{mescheder19cvpr,xie19iccv,choy16eccv,wang18eccv}.
We start with an exact comparison to ONN~\cite{mescheder19cvpr}. For this we
use the open source implementation provided by the authors to train and test
their model on our low-realism images train and test sets. We then use our
evaluation procedure on their output predictions.
As table~\ref{tab:single-object-star} shows, ONN achieves $52.6\%$ mIoU on our
data with our evaluation (and 51.5\% with ONN's evaluation procedure). This
number is expectedly lower than the $57.1\%$ reported
in~\cite{mescheder19cvpr} as we ask ONN to reconstruct each shape at the pose
depicted in the input image, instead of the canonical pose. From ONN's
perspective, the training set contains 24 times more different shapes, one for
each rendered view of an object.
Our best model for low-realism renderings $h_5$ outperforms ONN on every class
and achieves $57.9\%$ mIoU. ONN's performance is comparable to $h_2$, our best
model that, like ONN, does not use skip connections.

\begin{table}[t]
  \centering
  \fontsize{9}{11}
  \selectfont
  \begin{tabular}{c|c|ccccccccccccc}
  \textbf{model} & \rott{\textbf{mIoU}} & \rott{airplane} & \rott{bench} & \rott{cabinet} & \rott{car}  & \rott{chair} & \rott{display} & \rott{lamp} & \rott{loudspeaker} & \rott{rifle} & \rott{sofa} & \rott{table} & \rott{telephone} & \rott{vessel} \\
  \hline
  $ONN^*$   & 52.6 & 45.8 & 45.1 & 43.8 & 54.0 & 58.5 & 55.4 & 39.5 & 57.0 & 48.0 & 68.0 & 50.7 & 68.3 & 49.9 \\
  $h_2$     & 53.0 & 46.9 & 44.3 & 44.7 & 56.4 & 57.4 & 53.8 & 35.9 & 58.1 & 53.4 & 67.2 & 49.7 & 70.9 & 49.9 \\
  $h_5$     & 57.9 & 53.0 & 50.8 & 50.9 & 57.3 & 63.0 & 57.2 & 42.1 & 60.8 & 64.6 & 70.6 & 55.5 & 73.1 & 54.0 \\
  \hline
  3D-R2N2   & 49.3 & 42.6 & 37.3 & 66.7 & 66.1 & 43.9 & 44.0 & 28.1 & 61.1 & 37.5 & 62.6 & 42.0 & 61.1 & 48.2 \\
  Pix2Mesh  & 48.0 & 42.0 & 32.3 & 66.4 & 55.2 & 39.6 & 49.0 & 32.3 & 59.9 & 40.2 & 61.3 & 39.5 & 66.1 & 39.7 \\
  ONN       & 57.1 & 57.1 & 48.5 & 73.3 & 73.7 & 50.1 & 47.1 & 37.1 & 64.7 & 47.4 & 68.0 & 50.6 & 72.0 & 53.0 \\
  \end{tabular}
  \vspace{1.0em}
  \caption{
  \small
  \smallcap
  Comparison to state of the art. The first three rows compare
  ONN~\cite{mescheder19cvpr} to our models $h_2$ and $h_5$, all trained on our
  data.
  The next three rows are taken from~\cite{mescheder19cvpr} and report
  performance of 3D-R2N2~\cite{choy16eccv},
  Pix2Mesh~\cite{wang18eccv}, and ONN~\cite{mescheder19cvpr} on their data.
  }
  \label{tab:single-object-star}
  \vspace{-2em}
\end{table}

We then compare to 3D-R2N2~\cite{choy16eccv}, Pix2Mesh~\cite{wang18eccv}, and
again ONN~\cite{mescheder19cvpr}, using their mIoU as reported
by~\cite{mescheder19cvpr} (table~\ref{tab:single-object-star}). Our model
$h_5$ clearly outperforms 3D-R2N2 (+$8.6\%$) and Pix2Mesh (+$9.9\%$). It also
reaches a slightly better mIoU than ONN (+$0.8\%$), while reconstructing in
the appropriate pose for each input image, as opposed to a fixed canonical
pose.
We also compare on the Chamfer Distance surface metric, implemented exactly as
in~\cite{mescheder19cvpr}. We obtain $0.15$, which is better than 3D-R2N2
(0.278), Pix2Mesh (0.216), and ONN (0.215), all compared with the same metric
(as reported by~\cite{mescheder19cvpr}).%
\footnote{Several other works~\cite{fan17cvpr,groueix18cvpr}, including very
recent ones~\cite{chen20cvpr,yao20cvpr}, report Chamfer Distance and not IoU.
They adopt subtly different implementations, varying the underlying point
distance metric, scaling, point sampling, and aggregation across points. Thus,
they report different numbers for the same works, preventing direct comparison.
}

Finally, we compare to Pix2Vox~\cite{xie19iccv} and its extension
Pix2Vox++~\cite{xie20ijcv} (concurrent work to ours). For a fair comparison we
evaluate our $h_5$ model on a $32^3$ grid of points, matching the $32^3$ voxel
grid output by~\cite{xie19iccv,xie20ijcv}. We compare directly to the mIoU they
report. Our model $h_5$ achieves 68.9\% mIoU in this case, $+2.8\%$ higher than
Pix2Vox (66.1\% for their best Pix2Vox-A model) and $+1.9\%$ higher than
Pix2Vox++ (67.0\% for their best Pix2Vox++/A model).

\mypar{Reconstructing at high resolutions.}
Our model can perform reconstruction at a higher resolution than the one used
during training (sec.~\ref{sect:3d-volume-representation}). We study this here
by reconstructing at $2\times$ and $4\times$ higher resolution.
We train one model ($y_1$) using a $32^3$ grid and one ($y_2$) using a $64^3$
grid, with ray-traced skip connections, images with low realism, and focal
loss. We then reconstruct a $128^3$ discretization from each model, by running
inference multiple times at different grid offsets (64 and 8 times,
respectively, sec.~\ref{sect:mesh-reconstruction}). At test time, we always
measure performance on the $128^3$ reconstruction, regardless of training
resolution.

Fig.~\ref{fig:superres-results} shows example reconstructions. We compare
performance to $h_4$ from table~\ref{tab:single-object-hyper-param}, which was
trained with same settings but at native grid resolution $128^3$. Our first
model (trained on $32^3$ and producing $128^3$) achieves $53.1\%$ mIoU. The
second model (trained on $64^3$ and producing $128^3$) gets to $56.1\%$,
comparable to $h_4$ ($56.6\%$). This demonstrates that we can reconstruct at
substantially higher resolution than the one used during training.

\subsection{Single object reconstruction on Pix3D}
\label{sect:pix3d-results}

We evaluate the performance of our method on real images using the Pix3D
dataset~\cite{sun18cvpr}, which contains 10069 images annotated with 395 unique
3D models from 9 classes (bed, bookcase, chair, desk, misc, sofa, table, tool,
wardrobe). Most of the images are of indoor scenes, with complex backgrounds,
occlusion, shadows, and specular highlights.

The images often contain multiple objects, but Pix3D provides annotations for
exactly one of them per image. To deal with this discrepancy, we use a single
object reconstruction pipeline.
At test time, our method looks at the whole image, but we only reconstruct the
volume inside the 3D box of the object annotated in the ground-truth. This is
similar to how other methods deal with this discrepancy at test time\footnote{
Pix2Vox~\cite{xie19iccv} and Pix2Vox++~\cite{xie20ijcv} crop the input image
before reconstruction, using the 2D projected box of the ground-truth object.
MeshRCNN~\cite{gkioxari19iccv} requires the ground-truth object 3D center as
input. It also crops the image through the ROI pooling layers, using the 2D
projected ground-truth box to reject detections with IoU $<0.3$.}.

\begin{figure}[t!]
  \centering
  \small
  \pixv{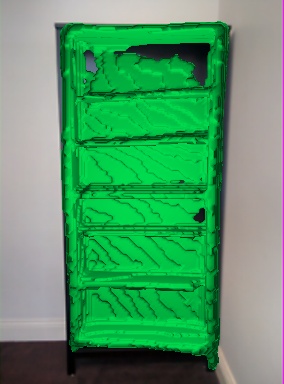}{0}{2}{33mm}
  \pixh{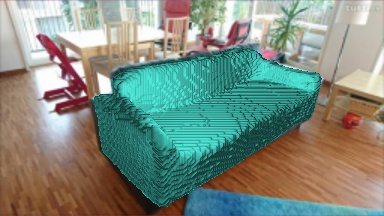}{1}{2}{42mm}
  \pixv{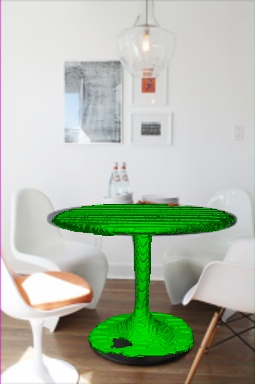}{2}{3}{33mm}\\[2mm]
  \pixh{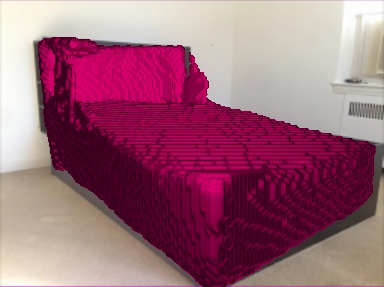}{2}{1}{33mm}
  \pixv{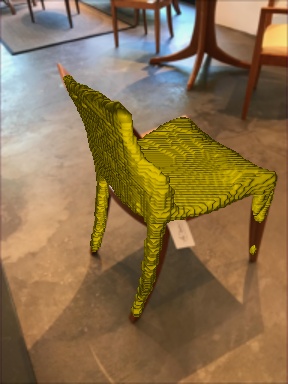}{1}{3}{33mm}
  \pixh{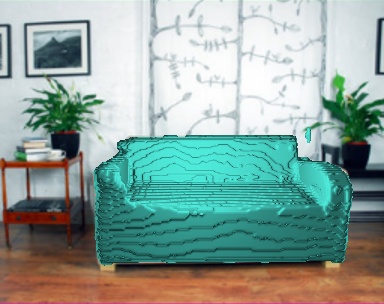}{3}{2}{30mm}

  \caption{
  \small
  \smallcap
  Qualitative results on Pix3D. For each example, the large image shows our
  reconstruction overlaid on the RGB input. The smaller images show the RGB
  input, and our reconstruction viewed from two additional viewpoints.}
  \label{fig:pix3d-result-images}
  \vspace{-1.8em}
\end{figure}

\mypar{Generalization across domains (ShapeNet to Pix3D).}
We first perform experiments in the same settings as previous
works~\cite{choy16eccv,sun18cvpr,xie19iccv}, which train on synthetic images of
ShapeNet objects. As they do, we focus on chairs. We train on the high-realism
synthetic images from sec.~\ref{sect:single-object-performance}. For each image
we crop out the chair and paste it over a random background from
OpenImages~\cite{kuznetsova18arxiv,openimages}, a random background from
SunDB~\cite{xiao10cvpr}, and a white background. We start from a model
pre-trained on ShapeNet ($h_7$, sec.~\ref{sect:single-object-performance}) and
continue training on this data.

We evaluate on the 2894 Pix3D images with chairs that are neither occluded nor
truncated. We predict occupancy on a $32^3$ discretization of 3D space. This is
the exact same setting used in~\cite{choy16eccv,sun18cvpr,xie19iccv,xie20ijcv}.
Our model achieves 29.7\% IoU, which is higher than Pix2Vox~\cite{xie19iccv}
($28.8\%$, for their best Pix2Vox-A model), the Pix3D method~\cite{sun18cvpr}
($28.2\%$, for their best `with pose' model), 3D-R2N2~\cite{choy16eccv}
($13.6\%$, as reported in~\cite{sun18cvpr}), and the concurrent work
Pix2Vox++~\cite{xie20ijcv} ($29.2\%$ for their best Pix2Vox++/A model).

This setting is motivated by the fact that most real-world images do not come
with annotations for the ground-truth 3D shape of the objects in them.
Therefore, it represents the common scenario of training from synthetic data
with available 3D supervision.

\mypar{Fine tuning on real data from Pix3D.}
We now consider the case where we do have access to a small set of real-world
images with ground-truth 3D shapes for training. For this we use the $S_1$ and
$S_2$ train/test splits of Pix3D defined in~\cite{gkioxari19iccv}. There are no
images in common between the test and train splits in both $S_1$ and $S_2$.
Furthermore, in $S_2$ also the set of object instances is disjoint between train
and test splits. In $S_1$ instead, some objects are allowed to be in both the
splits, albeit with a different pose and against a different
background.

We train two models, one for $S_1$ and one for $S_2$. In both cases, we start
from a model pre-trained on ShapeNet ($h_7$) and we then continue training on
the respective Pix3D train set.
On average over all 9 object classes, we achieve $33.3\%$ mIoU on the test set
of $S_1$, and $23.6\%$ on the test set of $S_2$, when evaluating at $128^3$
discretization of 3D space (fig.~\ref{fig:pix3d-result-images}).

As a reference, we compare to a model trained only on ShapeNet. As above, we
start from $h_7$ and we augment with real-world backgrounds. We evaluate
performance on all 9 object classes on the test splits of $S_1$ and $S_2$. This
leads to $20.9\%$ mIoU for S1 and $20.0\%$ for $S_2$. This confirms that
fine-tuning on real-world data from Pix3D performs better than training purely
on synthetic data.

\begin{figure}[t!]
  \centering
  \includegraphics[width=.135\linewidth]{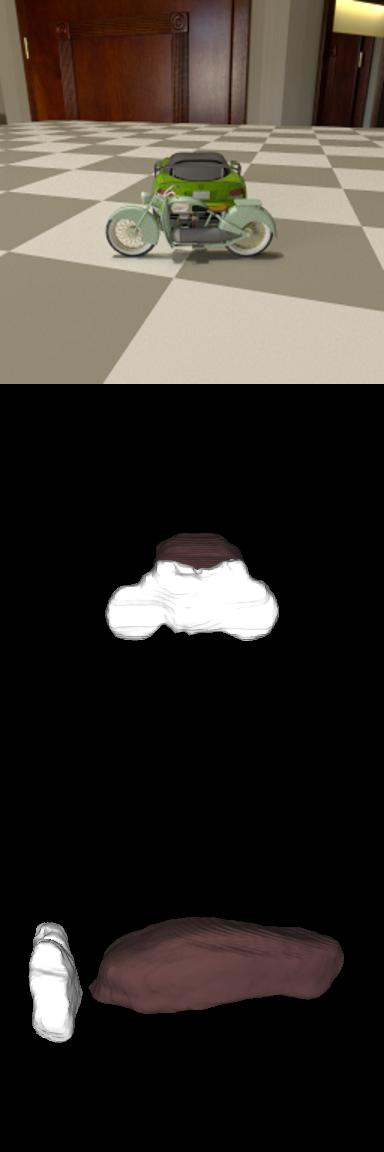}
  \includegraphics[width=.135\linewidth]{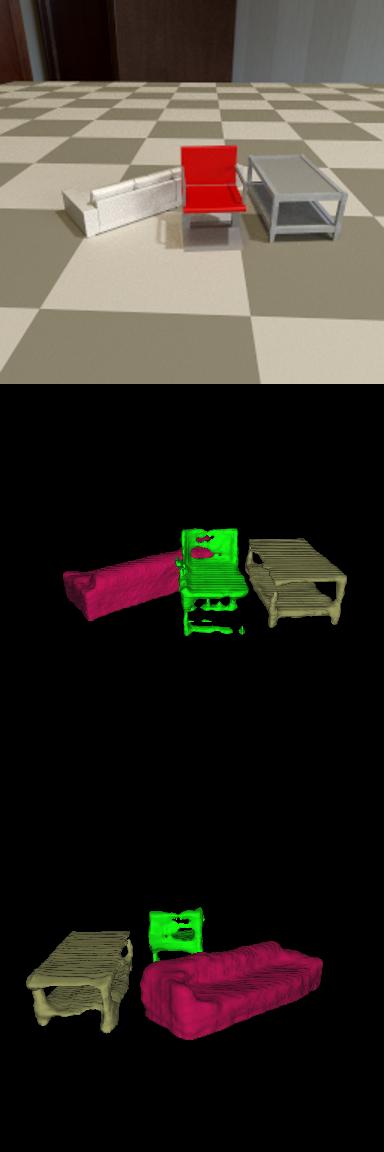}
  \includegraphics[width=.135\linewidth]{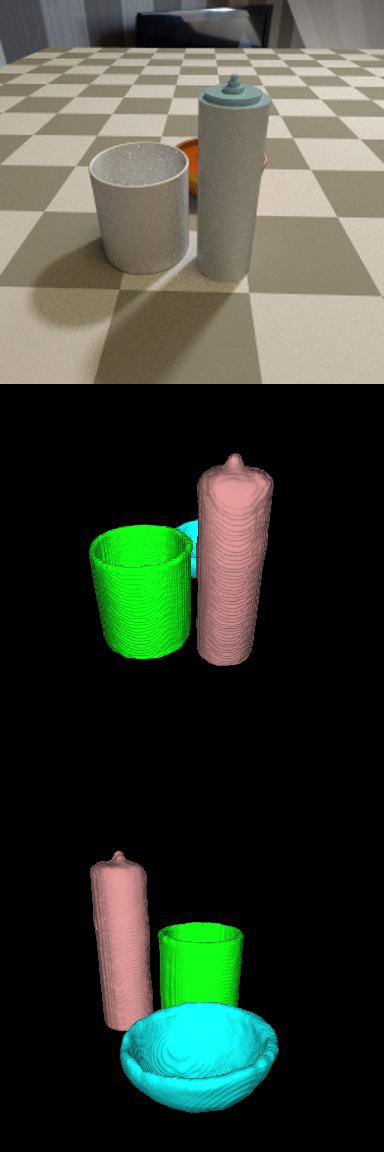}
  \includegraphics[width=.135\linewidth]{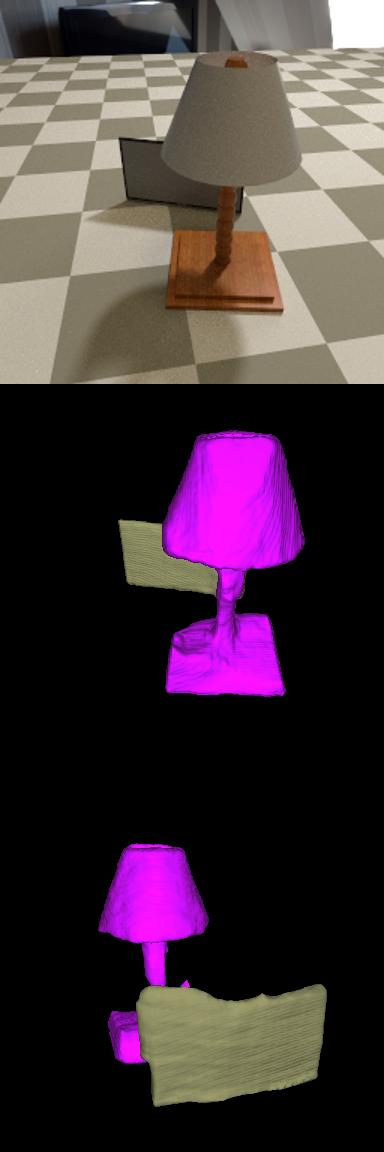}
  \includegraphics[width=.135\linewidth]{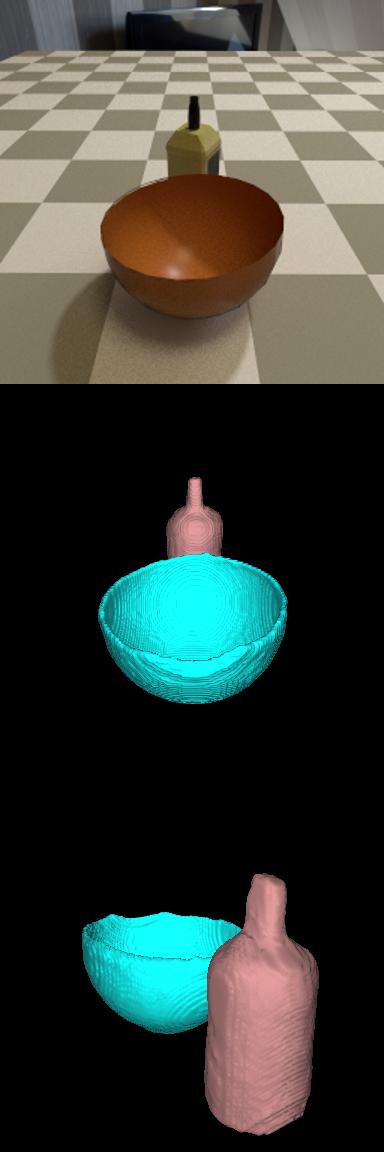}
  \includegraphics[width=.135\linewidth]{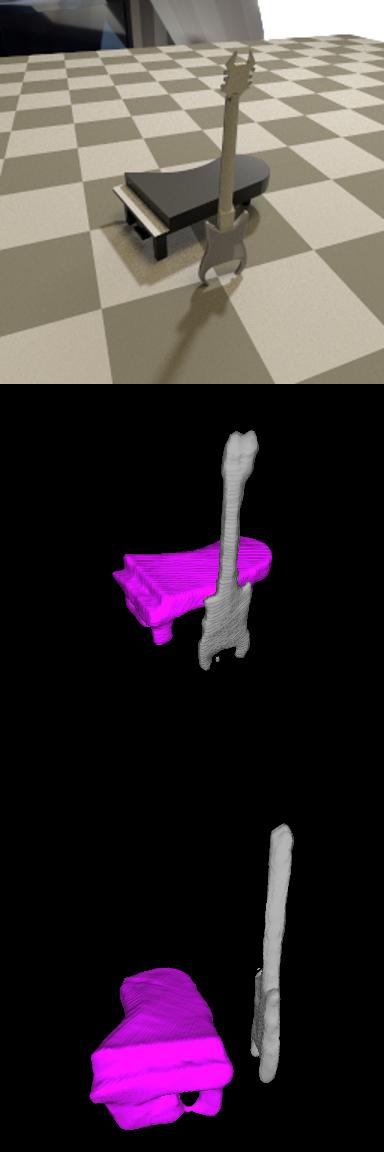}
  \includegraphics[width=.135\linewidth]{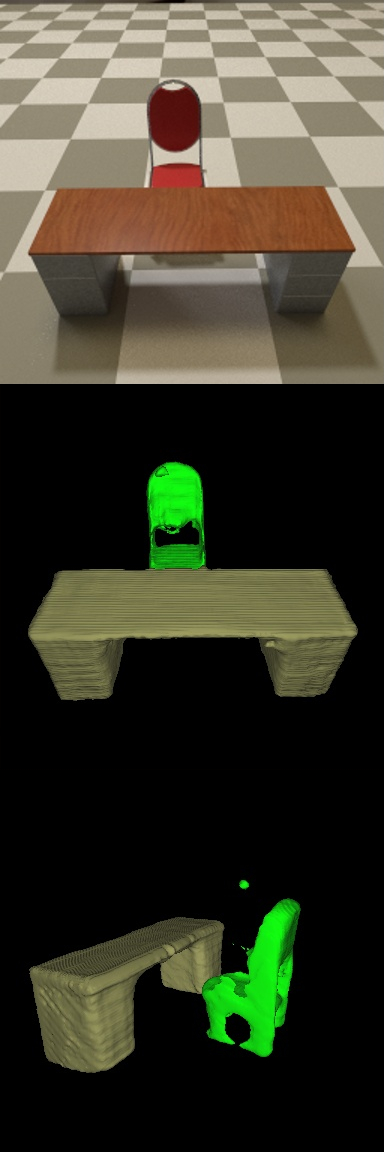}
  \vspace{-0.5em}
  \caption{
  \small
  \smallcap
  Pairs and triplets reconstructed by $m_7$ and $m_9$,
  shown from the camera and from one additional viewpoint. Our model
  hallucinates the occluded parts and reconstructs all objects
  in their correct spatial arrangement, in a common coordinate frame. }
  \label{fig:multi-obj-reconstr}
  \vspace{-1.5em}
\end{figure}

\vspace{2em}
\subsection{Multiple object reconstruction}
\label{sect:multiclass-reconstruction}

\mypar{Datasets and settings.}
We construct two datasets by assembling objects from ShapeNet.
The first is {\em ShapeNet-pairs}, with
several pairs of object classes: bed-pillow, bottle-bowl, bottle-mug,
chair-table, display-lamp, guitar-piano, motorcycle-car. The second is {\em
ShapeNet-triplets}, with bottle-bowl-mug and chair-sofa-table.
We randomly sample the object instances participating in each combination from
ShapeNet, respecting its official trainval and test splits. For each image
we generate, we random sample two/three object instances, place them at random
locations on the ground plane, with random scale and rotation, making sure they
do not overlap in 3D, and render the scene from a random camera viewpoint (yaw
and pitch). We construct the same number of scenes for every pair/triplet for
training and testing. Note how the objects' scales and rotations, as well as the
camera viewpoints, vary between the train and test splits and between images
within a split (but their overall distribution is the same). Like the
single-object case, the object instances are disjoint in the training and test
splits.
In total, for pairs we generate 365'600 images on trainval and 91'200 on test;
for triplets we make 91'400 on trainval and 22'000 on test.

We perform experiments varying the use of ray-traced skip connections, the image
realism, and the loss. Besides categorical cross entropy ($\textit{Xent}$) and
focal loss, we also combine Xent and $\ioug$ \eqref{eq:ioug} into a product. The
IoU part pushes the model to reconstruct full shapes, while Xent pushes it to
learn the correct class for each 3D point. We train on the train and val splits
together, and test on the test split, always with grid resolution $128^3$.

\mypar{Reconstruction performance.}
Table~\ref{tab:multi-object-results}(b) summarizes our results, including also
multi-class reconstruction of images showing a single ShapeNet object for
reference (bottom row, marked `single'). We show example reconstructions in
fig.~\ref{fig:multi-obj-reconstr}. On ShapeNet-pairs, using ray-traced skip
connections improves mIoU substantially (by $8-10\%$), in conjunction with
any loss function.
The improvement is twice as large than in the single object case
(table~\ref{tab:multi-object-results}), confirming that ray-traced skip
connections indeed help more for multiple objects. They allow the model to
propagate occlusion boundaries and object contact points detected on the 2D
image into 3D, and also to understand the depth relations among objects
locally.
When using skip connections, our IoU loss performs best, followed closely by the
focal loss. The cross-entropy loss underperforms in comparison ($-13\%$ mIoU).
As with the single-object case, results are slightly better on higher image
realism.

Importantly, we note that performance for pairs/triplets is only mildly lower
than for the easier single-object scenario. To investigate why, we compare the
single-object models $m_{11}$ and $h_7$ from
table~\ref{tab:multi-object-results}. They differ in the number of classes they
handle (14 for $m_{11}$, 2 for $h_7$) but have otherwise identical
settings. While the difference in their mean IoUs is $12\%$ ($46.9\%$ \vs
$59.1\%$), their global IoUs are close ($56.4\%$ \vs $59.3\%$). Hence, our model
is still good at reconstructing the overall shapes of objects, but makes some
mistakes in assigning the right class.

Finally, we note that reconstruction is slightly better overall for triplets
rather than for pairs. This is due to the different classes involved. On pairs
composed of the same classes appearing in the triplets, results are better for
pairs.

In conclusion, these results confirm that we are able to perform 3D
reconstruction in the harder multiple object scenario.

\begin{wrapfigure}{r}{0.35\linewidth}
 \vspace{-2.2em}
  \centering
  \includegraphics[width=.97\linewidth]{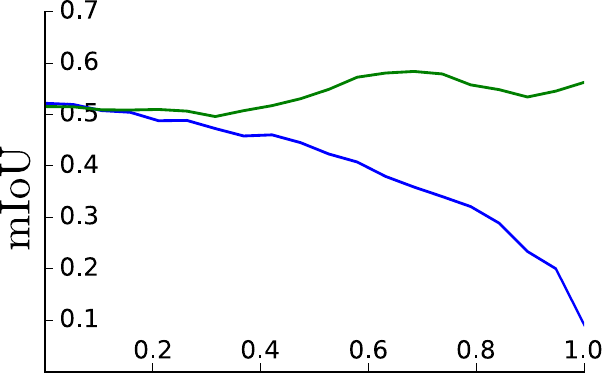}
 \vspace{-.5em}
  \caption{
  \small
  \smallcap
  mIoU \vs object \textcolor{blue}{occlusion} and
  \textcolor{OliveGreen}{depth}.}
  \label{fig:miou-vs-occlusion}
 \vspace{-2em}
\end{wrapfigure}

\mypar{Occlusion and distance.}
In fig.~\ref{fig:miou-vs-occlusion}, we break down the
performance (mIoU) of $m_7$ by the degree of object occlusion (blue),
and also by the object depth for unoccluded objects (i.e. distance to the
camera, green). The performance gracefully degrades as
occlusion increases, showing that our model can handle it well. Interestingly,
the performance remains steady with increasing depth, which correlates to
object size in the image. This shows that our model reconstructs far-away
objects about as well as nearby ones.

\mypar{Generalizations.}
In the suppl. material we explore even more challenging scenarios, where the
number of objects varies between training and test images, and where the test
set contains combintations of classes not seen during training.

\vspace{-.4em}
\section{Conclusions}

We made three contributions to methods for reconstructing the shape of a
single object given one RBG image as input:
(1) ray-traced skip connections that propagate local 2D information to the
output 3D volume in a physically correct manner;
(2) a hybrid 3D volume representation that enables building translation
equivariant models, while at the same time producing fine object details
with limited memory;
(3) a reconstruction loss tailored to capture overall object geometry.
We then adapted our model to reconstruct multiple objects.
By doing so jointly in a single pass, we produce a coherent
reconstruction with all objects in one consistent 3D coordinate
frame, and without intersecting in 3D space.
Finally, we validated the impact of our contributions on synthetic data from
ShapeNet as well as real images from Pix3D, including a full quantitative
evaluation of 3D shape reconstruction of multiple objects in the same image.

{\small
\bibliographystyle{splncs04}
\bibliography{longstrings,shortstrings,corenet}
}

\end{document}